\documentclass{tlp}

\usepackage{amsmath}
\usepackage{graphicx}
\usepackage{multirow}
\usepackage{hyperref}
\usepackage{url}
\usepackage{float}
\usepackage{caption}
\usepackage{subcaption}
\usepackage{booktabs}
\usepackage{fixltx2e}
\usepackage[style=authoryear]{biblatex}
\begin{document}

\lefttitle{Hillerström and Burghouts}

\jnlPage{1}{16}
\jnlDoiYr{2024}
\doival{10.1017/xxxxx}

\title[Probabilistic Inductive Logic Programming]{Towards Probabilistic Inductive Logic Programming with Neurosymbolic Inference and Relaxation}

\begin{authgrp}
\author{\gn{F. Hillerström and G.J. Burghouts}}
\affiliation{TNO, The Netherlands}
\end{authgrp}


\maketitle

\begin{abstract}
Many inductive logic programming (ILP) methods are incapable of learning programs from probabilistic background knowledge, e.g. coming from sensory data or neural networks with probabilities. We propose Propper, which handles flawed and probabilistic background knowledge by extending ILP with a combination of neurosymbolic inference, a continuous criterion for hypothesis selection (BCE) and a relaxation of the hypothesis constrainer (NoisyCombo). For relational patterns in noisy images, Propper can learn programs from as few as 8 examples. It outperforms binary ILP and statistical models such as a Graph Neural Network.
\end{abstract}


\section{Introduction}

Inductive logic programming (ILP) \autocite{muggleton1995inverse} learns a logic program from labeled examples and background knowledge (e.g. relations between entities). Due to the strong inductive bias imposed by the background knowledge, ILP methods can generalize from small numbers of examples \autocite{cropper2022inductive}. Other advantages are the ability to learn complex relations between the entities, the expressiveness of first-order logic, and the resulting program can be understood and transferred easily because it is in symbolic form \autocite{cropper2022_30newintro}. This makes ILP an attractive alternative methodology besides statistical learning methods.

For many real-world applications, dealing with noise is essential. Mislabeled samples are one source of noise. To learn from noisy labels, various ILP methods have been proposed to generalize a subset of the samples \autocite{srinivasan2001aleph,ahlgren2013efficient,zeng2014quickfoil,raedt2015inducing}. To advance methods to learn recursive programs and invent new predicates, Combo \autocite{cropper2023learning} was proposed, a method that searches for small programs that generalize subsets of the samples and combines them. MaxSynth \autocite{hocquette2024learning} extends Combo to allow for mislabeled samples, while trading off program complexity for training accuracy. These methods are dealing with noisy labels, but do not explicitly take into account errors in the background knowledge, nor are they designed to deal with probabilistic background knowledge. 

Most ILP methods take as a starting point the inputs in symbolic declarative form \autocite{cropper2021turning}. Real-world data often does not come in such a form. A predicate $p(.)$, detected in real-world data, is neither binary or perfect. The assessment of the predicate can be uncertain, resulting in a non-binary, probabilistic predicate. Or the assessment can be wrong, leading to imperfect predicates. Dealing with noisy and probabilistic background knowledge is relevant for learning from sources that exhibit uncertainties. A probabilistic source can be a human who needs to make judgements at an indicated level of confidence. A source can also be a sensor measurement with some confidence. For example, an image is described by the objects that are detected in it, by a deep learning model. Such a model predicts locations in the image where objects may be, at some level of confidence. Some objects are detected with a lower confidence than others, e.g. if the object is partially observable or lacks distinctive visual features. The deep learning model implements a probabilistic predicate that a particular image region may contain a particular object, e.g. 0.7 :: vehicle(x). Given that most object detection models are imperfect in practice, it is impossible to determine a threshold that distinguishes the correct and incorrect detections. 

Two common ILP frameworks, Aleph \autocite{srinivasan2001aleph} and Popper \autocite{learning_from_failures}, typically fail to find the correct programs when dealing with predicted objects in images \autocite{helff2023v}; even with a state-of-the-art object detection model, and after advanced preprocessing of said detections. In the absence of an ideal binarization of probabilities, most ILP methods are not applicable to probabilistic sources \autocite{cropper2021turning}.

\begin{figure}[H]
  \includegraphics[scale=.293]{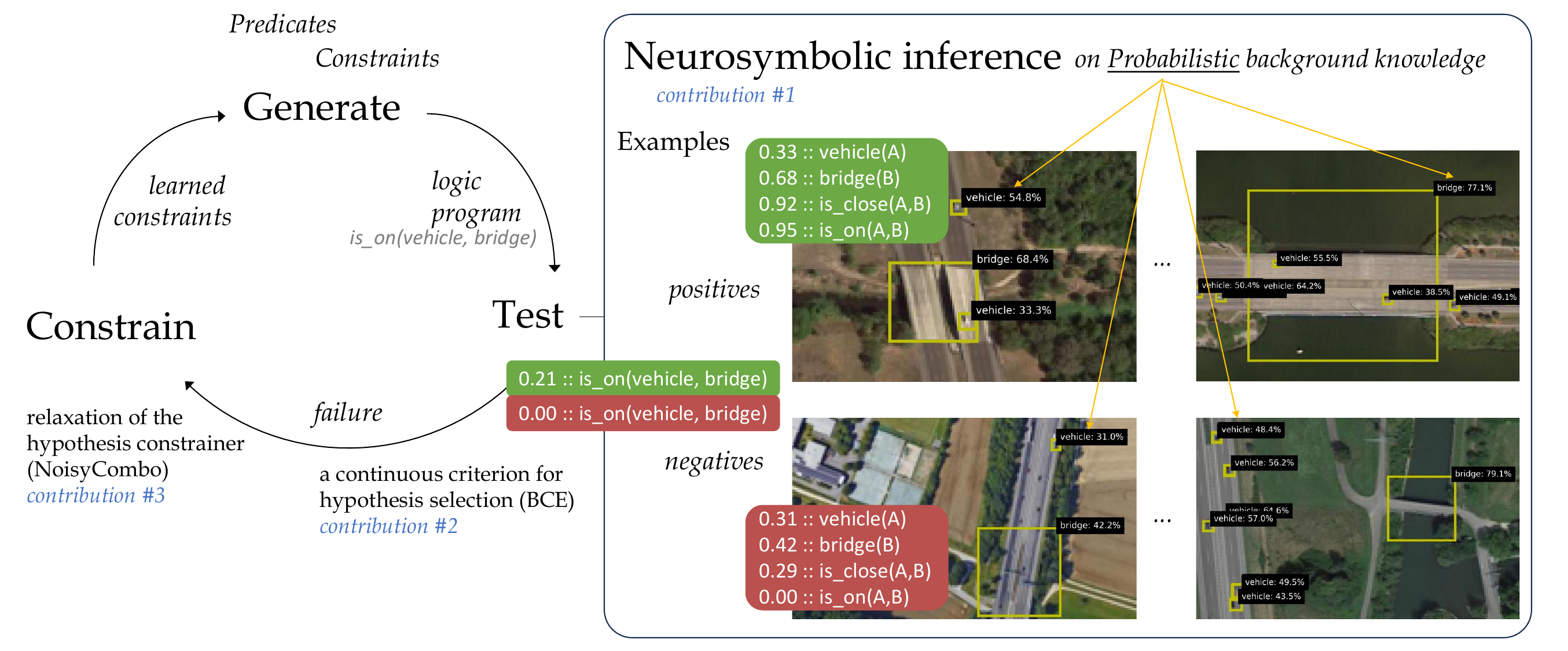}
  \caption{Our method Propper extends the ILP method Popper that learns from failures (left) with neurosymbolic inference to test logical programs on probabilistic background knowledge, e.g. objects detected in images with a certain probability (right).}
  \label{method}
\end{figure}

We propose a method towards probabilistic ILP. At a high level, ILP methods typically induce a logical program that entails many positive and few negative samples, by searching the hypothesis space, and subsequently testing how well the current hypothesis fits the training samples \autocite{cropper2022_30newintro}. One such method is Popper, which learns from failures (LFF) \autocite{learning_from_failures}, in an iterative cycle of generating hypotheses, testing them and constraining the hypothesis search. Our proposal is to introduce a probabilistic extension to LFF at the level of hypothesis testing. For that purpose, we consider neurosymbolic AI \autocite{hybrid_ai}. Within neurosymbolic AI a neural network predicts the probability for a predicate. For example a neural network for object detection, which outputs a probability for a particular object being present in an image region, e.g., 0.7 :: vehicle(x). Neurosymbolic AI connects this neural network with knowledge represented in a symbolic form, to perform reasoning over the probabilistic predicates predicted by the neural network. With this combination of a neural network and symbolic reasoning, neurosymbolic AI can reason over unstructured inputs, such as images. We leverage neurosymbolic programming and connect it to the tester within the hypothesis search. One strength of neurosymbolic programming is that it can deal with uncertainty and imperfect information \autocite{hybrid_ai,neuro_symbolic,scallop,scallop_foundationmodels}, in our case the probablistic background knowledge. 

We propose to use neurosymbolic inference as tester in the test-phase of the LFF cycle. Neurosymbolic reasoning calculates an output probability for a logical query being true, for every input sample. The input samples are the set of positive and negative examples, together with their probabilistic background knowledge. The logical query evaluated within the neurosymbolic reasoning is the hypothesis generated in the generate-phase of the LFF cycle, which is a first-order-logic program. With the predicted probability of the hypothesis being true per sample, it becomes possible to compute how well the hypothesis fits the training samples. That is used to continue the LFF cycle and generate new constraints based on the failures. 


Our contribution is a step towards probabilistic ILP by proposing a method called Propper. It builds on an ILP framework that is already equipped to deal with noisy labels, Popper-MaxSynth \autocite{learning_from_failures,hocquette2024learning}, which we extend with neurosymbolic inference which is able to process probabilistic facts, i.e. uncertain and imperfect background knowledge. Our additional contributions are a continuous criterion for hypothesis selection, that can deal with probabilities, and a relaxed formulation for constraining the hypothesis space. Propper and the three contributions are outlined in Figure \ref{method}. We compare Popper and Propper with statistical ML models (SVM and Graph Neural Network) for the real-life task of finding relational patterns in satellite images based on objects predicted by an imperfect deep learning model. We validate the learning robustness and efficiency of the various models. We analyze the learned logical programs and discuss the cases which are hard to predict. 

\section{Related Work}

For the interpretation of images based on imperfect object predictions, ILP methods such as Aleph \autocite{srinivasan2001aleph} and Popper \autocite{learning_from_failures} proved to be vulnerable and lead to incorrect programs or not returning a program at all \autocite{helff2023v}. Solutions to handle observational noise were proposed \autocite{cropper2021beyondentailment} for small binary images. With LogVis \autocite{muggleton2018meta} images are analyzed via physical properties. This method could estimate the direction of the light source or the position of a ball from images in very specific conditions or without clutter or distractors. $Meta_{Abd}$ \autocite{dai2020abductive} jointly learns a neural network with induction of recursive first-order logic theories with predicate invention. This was demonstrated on small binary images of digits. Real-life images are more complex and cluttered. We aim to extend these works to realistic samples, e.g. large color images that contain many objects under partial visiblity and in the midst of clutter, causing uncertainties. Contrary to $Meta_{Abd}$, we take pretrained models as a starting point, as they are often already very good at their task of analyzing images. Our focus is on extending ILP to handle probabilistic background knowledge.

In statistical relational artificial intelligence (StarAI) \autocite{raedt2016statistical} the rationale is to directly integrate probabilities into logical models. StarAI addresses a different learning task than ILP: it learns the probabilistic parameters of a given program, whereas ILP learns the program \autocite{cropper2021turning}. Probabilities have been integrated into ILP previously. Aleph \autocite{srinivasan2001aleph} was used to find interesting clauses and then learn the corresponding weights \autocite{huynh2008discriminative}. ProbFOIL \autocite{raedt2015inducing} and SLIPCOVER \autocite{bellodi2015structure} search for programs with probabilities associated to the clauses, to deal with the probabilistic nature of the background knowledge. SLIPCOVER searches the space of probabilistic clauses using beam search. The clauses come from Progol \autocite{muggleton1995inverse}. Theories are searched using greedy search, where refinement is achieved by adding a clauses for a target predicate. As guidance the log likelihood of the data is considered. SLIPCOVER operates in a probabilistic manner on binary background knowledge, where our goal is to involve the probabilities associated explicitly the background knowledge.

How to combine these probabilistic methods with recent ILP frameworks is unclear. In our view, it is not trivial and possibly incompatible. Our work focuses on integrating a probabilistic method into a modern ILP framework, in a simple yet elegant manner. We replace the binary hypothesis tester of Popper \autocite{learning_from_failures} by a neurosymbolic program that can operate on probabilistic and imperfect background knowledge \autocite{hybrid_ai,neuro_symbolic}. Rather than advanced learning of both the knowledge and the program, e.g. NS-CL \autocite{mao2019neuro}, we take the current program as the starting point. Instead of learning parameters, e.g. Scallop \autocite{scallop}, we use the neurosymbolic program for inference given the program and probabilistic background knowledge. Real-life samples may convey large amounts of background knowledge, e.g. images with many objects and relations between them. Therefore, scalability is essential. Scallop \autocite{scallop} improved the scalability over earlier neurosymbolic frameworks such as DeepProbLog \autocite{deepproblog,deepproblog_efficient}. Scallop introduced a tunable parameter $k$ to restrain the validation of hypotheses by analyzing the top-$k$ proofs. They asymptotically reduced the computational cost while providing relative accuracy guarantees. This is beneficial for our purpose. By replacing only the hypothesis tester, the strengths of ILP (i.e. hypothesis search) are combined with the strengths of neurosymbolic inference (i.e. probabilistic hypothesis testing).

\section{Propper Algorithm}
To allow ILP on flawed and probabilistic background knowledge, we extend modern ILP (Section \ref{sec_popper}) with neurosymbolic inference (\ref{sec_scallop}) and coin our method Propper. The neurosymbolic inference requires program conversion by grammar functions (\ref{sec_popper2scallop}), and we added a continuous criterion for hypothesis selection (\ref{sec_criterion}), and a relaxation of the hypothesis constrainer (\ref{sec_constrainer}).

\subsection{ILP: Popper}\label{sec_popper}

Popper represents the hypothesis space as a constraint satisfaction problem and generates constraints based on the performance of earlier tested hypotheses. It works by learning from failures (LFF) \autocite{learning_from_failures}. Given background knowledge $B$, represented as a logic program, positive examples $E^+$ and negative examples $E^-$, it searches for a hypothesis $H$ that is complete ($\forall e \in E^+, H \cup B \models e$) and consistent ($\forall e \in E^-, H \cup B \not\models e$). The algorithm consists of three main stages (see Figure \ref{method}, left). First a hypothesis in the form of a logical program is generated, given the known predicates and constraints on the hypothesis space. The Test stage tests the generated logical program against the provided background knowledge and examples, using Prolog for inference. It evaluates whether the examples are entailed by the logical program and background knowledge. From this information, failures that are made when applying the current hypothesis, can be identified. These failures are used to constrain the hypothesis space, by removing specializations or generalizations from the hypothesis space. In the original Popper implementation \autocite{learning_from_failures}, this cycle is repeated until an optimal solution is found; the smallest program that covers all positives and no negative examples\footnote{
See \autocite{learning_from_failures} for a formal definition.}. Its extension Combo combines small programs that do not entail any negative example \autocite{cropper2023learning}. When no optimal solution is found, Combo returns the obtained best solution. The Popper variant MaxSynth does allow noise in the examples and generates constraints based on a minimum description length cost function, by comparing the length of a hypothesis with the possible gain in wrongly classified examples \autocite{hocquette2024learning}.   

\subsection{Neurosymbolic Inference: Scallop} \label{sec_scallop} 

Scallop is a language for neurosymbolic programming which integrates deep learning with logical reasoning \autocite{scallop}. Scallop reasons over continuous, probabilistic inputs and results in a probabilistic output confidence. It consists of two parts: a neural model that outputs the confidence for a specific concept occurring in the data and a reasoning model that evaluates the probability for the query of interest being true, given the input. It uses provenance frameworks \autocite{kimmig2017algebraic} to approximate exact probabilistic inference, where the AND operator is evaluated as a multiplication ($AND(x, y) = x * y$), the OR as a minimization ($OR(x, y) = min(1, x + y)$) and the NOT as a $1-x$. Other, more advanced formulations are possible, e.g. $noisy$-$OR(x, y) = 1-(1-a)(1-b)$ for enhanced performance. For ease of integration, we considered this basic provenance. To improve the speed of the inference, only the most likely top-k hypotheses are processed, during the intermediate steps of computing the probabilities for the set of hypotheses.

\subsection{Connecting ILP and Neurosymbolic Inference} \label{sec_popper2scallop}
Propper changes the Test stage of the Popper algorithm (see Figure \ref{method}): the binary Prolog reasoner is replaced by the neurosymbolic inference using Scallop, operating on probabilistic background knowledge (instead of binary), yielding a probability for each sample given the logical program. The background knowledge is extended with a probability value before each first-order-logic statement, e.g. 0.7 :: vehicle(x). 

The Generate step yields a logic program in Prolog syntax. The program can cover multiple clauses, that can be understood as OR as one needs to be satisfied. Each clause is a function of predicates, with input arguments. The predicate arguments can differ between the clauses within the logic program. This is different from Scallop, where every clause in the logic program is assumed to be a function of the same set of arguments. As a consequence, the Prolog program requires syntax rewriting to arrive at an equivalent Scallop program. This rewriting involves three steps by consecutive grammar functions, which we illustrate with an example. Take the Prolog program:

\begin{align}
\begin{split}\label{eq_prolog}
    \texttt{f(A)} = {} & \texttt{has\_object(A, B), vehicle(B)}\\
    \texttt{f(A)} = {} & \texttt{has\_object(A, B), bridge(C), is\_on(B, C)}
\end{split}
\end{align}

The bodies of \texttt{f(A)} are extracted by: $b(\texttt{f})$ = \{[\texttt{has\_object(A, B), vehicle(B)}], [\texttt{has\_object(A, B), bridge(C), is\_on(B, C)}]\}. The sets of arguments of \texttt{f(A)} are extracted by: $v(\texttt{f}) = \{\{\texttt{A, B}\}, \{\texttt{C, A, B}\}\}$.

For a Scallop program, the clauses in the logic program need to be functions of the same argument set. Currently the sets are not the same: \{\texttt{A, B}\} vs. \{\texttt{C, A, B}\}. Function $e(\cdot)$ adds a dummy predicate for all non-used arguments, i.e. \texttt{C} in the first clause, such that all clauses operate on the same set, i.e. \{\texttt{C, A, B}\}:

\begin{align}
\begin{split}\label{eq_extend}
    e([\texttt{has\_object(A, B)}, {} & \texttt{vehicle(B)}], \{\texttt{C, A, B}\}) = \\ & \texttt{has\_object(A, B), vehicle(B), always\_true(C)}
\end{split}
\end{align}

After applying grammar functions $b(\cdot)$, $v(\cdot)$ and $e(\cdot)$, the Prolog program \texttt{f(A)} becomes the equivalent Scallop program \texttt{g(C, A, B)}:

\begin{align}
\begin{split}\label{eq_scallop}
    \texttt{g\textsubscript{0}(C, A, B)} = {} & \texttt{has\_object(A, B), vehicle(B), always\_true(C)}\\
    \texttt{g\textsubscript{1}(C, A, B)} = {} & \texttt{has\_object(A, B), bridge(C), is\_on(B, C)}\\
    \texttt{g(C, A, B)} = {} & \texttt{g\textsubscript{0}(C, A, B) or g\textsubscript{1}(C, A, B)}
\end{split}
\end{align}

\subsection{Selecting the Best Hypothesis} \label{sec_criterion}
MaxSynth uses a minimum-description-length (MDL) cost \autocite{hocquette2024learning} to select the best solution: 
\begin{equation}
    MDL_{B,E} = size(h) + fn_{B,E}(h) + fp_{B,E}(h)
\end{equation}
The MDL cost compares the number of correctly classified examples with the number of literals in the program. This makes the cost dependent on the dataset size and requires binary predictions in order to determine the number of correctly classified examples. Furthermore, it is doubtful whether the number of correctly classified examples can be compared directly with the rule size, since it makes the selection of the rule size dependent on the dataset size again. 

Propper uses the Binary Cross Entropy (BCE) loss to compare the performance of hypotheses, as it is a more continuous measure than MDL. The neurosymbolic inference predicts an output confidence for an example being entailed by the hypothesis. The BCE-cost compares this predicted confidence with the groundtruth (one or zero). For $y_i$ being the groundtruth label and $p_i$ the confidence predicted via neurosymbolic inference for example $i$, the BCE cost for $N$ examples becomes: 
\begin{equation}
    BCE = \frac{1}{N} \sum_{i=1}^{N}(y_i * log(p_i) + (1-y_i) * log(1-p_i)). 
\end{equation}

Scallop reasoning automatically avoids overfitting, by punishing the size of the program, because when adding more or longer clauses the probability becomes lower by design. The more ANDs in the program, the lower the output confidence of the Scallop reasoning, due to the multiplication of the probabilities. Therefore, making a program more specific will result in a higher BCE-cost, unless the specification is beneficial to remove FPs. Making the program more generic will cover more samples (due to the addition operator for the OR). However the confidences for the negative samples will increase as well, which will increase the BCE-cost again. The BCE-cost is purely calculated on the predictions itself, and thereby removes the dependency on the dataset size and the comparison between number of samples and program length.

\subsection{Constraining on Inferred Probabilities} \label{sec_constrainer}
Whereas Combo \autocite{cropper2023learning} and MaxSynth \autocite{hocquette2024learning} yield optimal programs given perfect background knowledge, with imperfect and probabilistic background knowledge no such guarantees can be provided. The probabilistic outputs of Scallop are converted into positives and negatives before constraining. The optimal threshold is chosen by testing 15 threshold values, evenly spaced between 0 and 1 and selecting the threshold resulting in the most highest true positives plus true negatives on the training samples. 

MaxSynth generates constraints based on the MDL loss \autocite{hocquette2024learning}, making the constraints dependent on the size of the dataset. To avoid this dependency, we introduce the NoisyCombo constrainer. Combo generates constraints once a false positive (FP) or negative (FN) is detected. $\exists e \in E^-, H \cup B \models e$: prune generalisations.
$\exists e \in E^+, H \cup B \not\models e$ or $\forall e \in E^-, H \cup B \not\models e$: prune specialisations. 
NoisyCombo relaxes this condition and allows a few FPs and FNs to exist, depending on an expected noise level, inspired by LogVis \autocite{muggleton2018meta}. This parameter defines a percentage of the examples that could be imperfect, from which the allowed number of FPs and FNs is calculated. $\sum(e \in E^-, H \cup B \models e) > noise\_level * N_{negatives}$: prune generalisations. $\forall e \in E^-, H \cup B \not\models e$: prune specialisations. The positives are not thresholded by the noise level, since programs that cover at least one positive sample are added to the combiner. 

\section{Analyses}

We validate Propper on a real-life task of finding relational patterns in satellite images, based on flawed and probabilistic background knowledge about the objects in the images, which are predicted by an imperfect deep learning model. We analyze the learning robustness under various degrees of flaws in the background knowledge. We do this for various models, including Popper (on which Propper is based) and statistical ML models. In addition, we establish the learning efficiency for very low amounts of training data, as ILP is expected to provide an advantage because it has the inductive bias of background knowledge. We analyze the learned logical programs, to compare them qualitatively against the target program. Finally, we discuss the cases that are hard to predict. 

\subsection{First Dataset}

The DOTA dataset \autocite{xia2018dota} contains many satellite images. This dataset is very challenging, because the objects are small, and therefore visual details are lacking. Moreover, some images are very cluttered by sometimes more than 100 objects. 

\begin{figure}[H]
    \centering
     \begin{subfigure}[b]{0.48\textwidth}
         \centering
         \includegraphics[width=\textwidth]{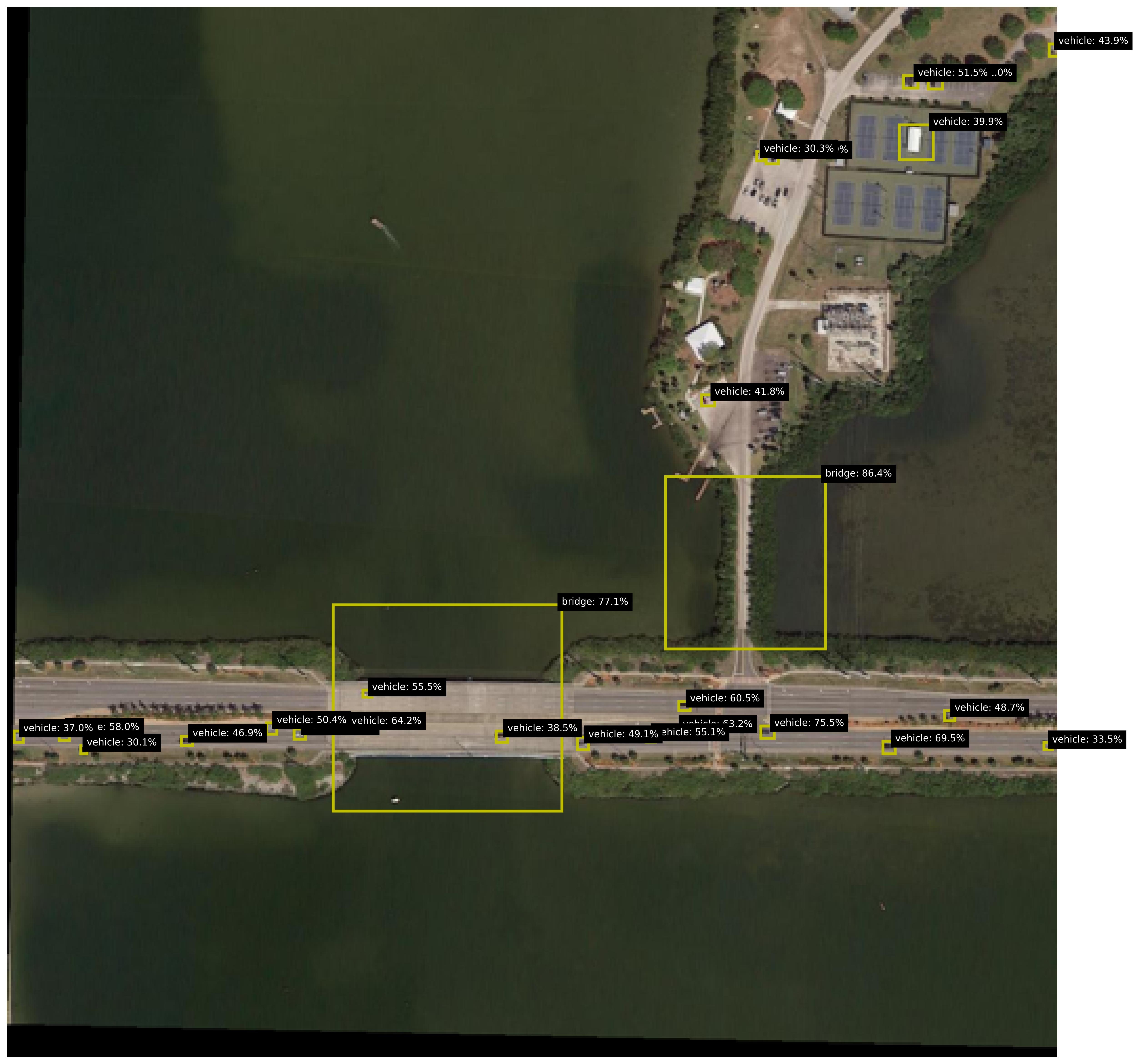}
         \caption{Positive image}
         \label{pos_img}
     \end{subfigure}
     \begin{subfigure}[b]{0.45\textwidth}
         \centering
         \includegraphics[width=\textwidth]{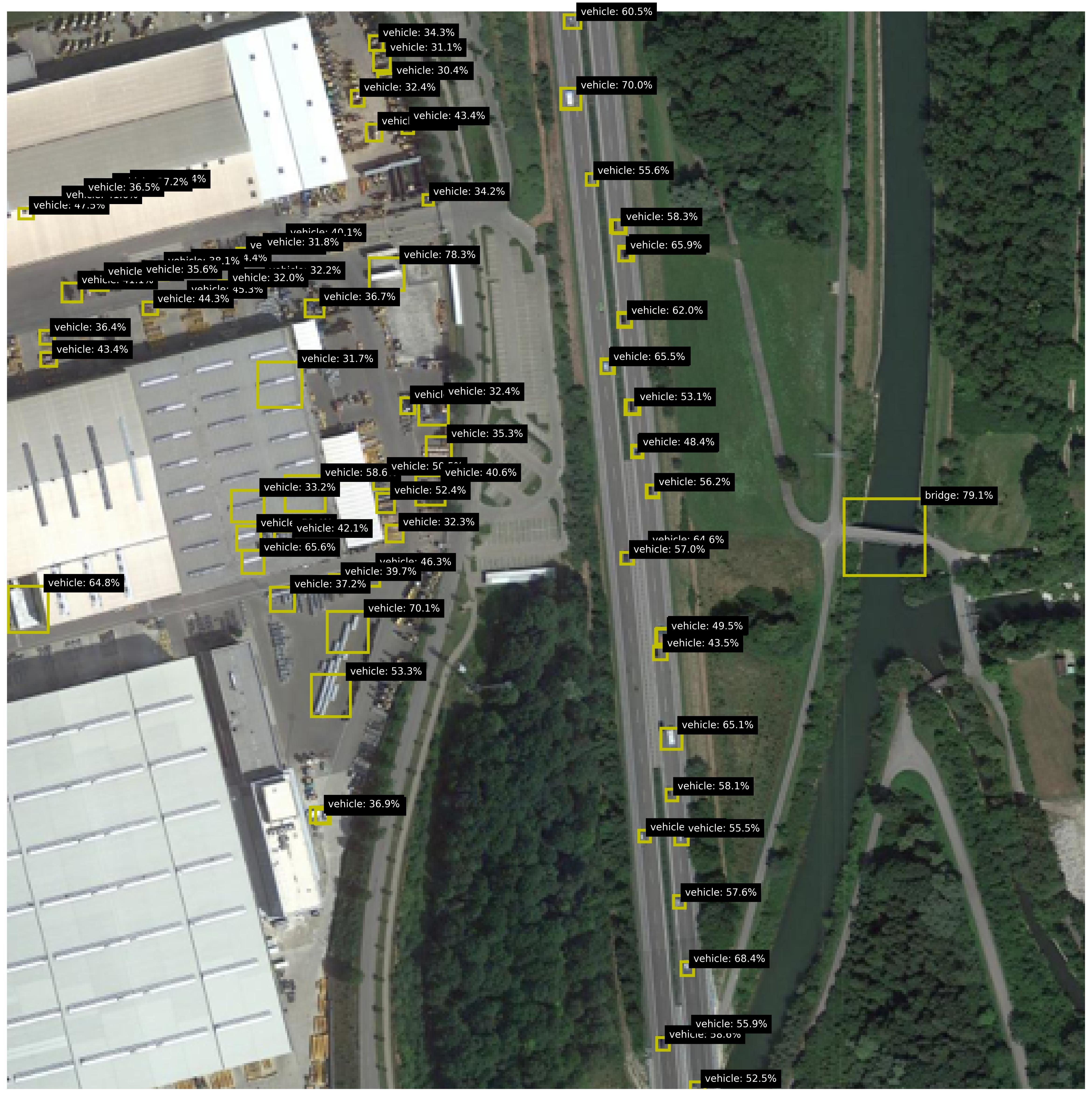}
         \caption{Negative image}
         \label{neg_img}
     \end{subfigure}
     \\
     \centering
     \begin{subfigure}[b]{0.435\textwidth}
         \centering
         \includegraphics[width=\textwidth]{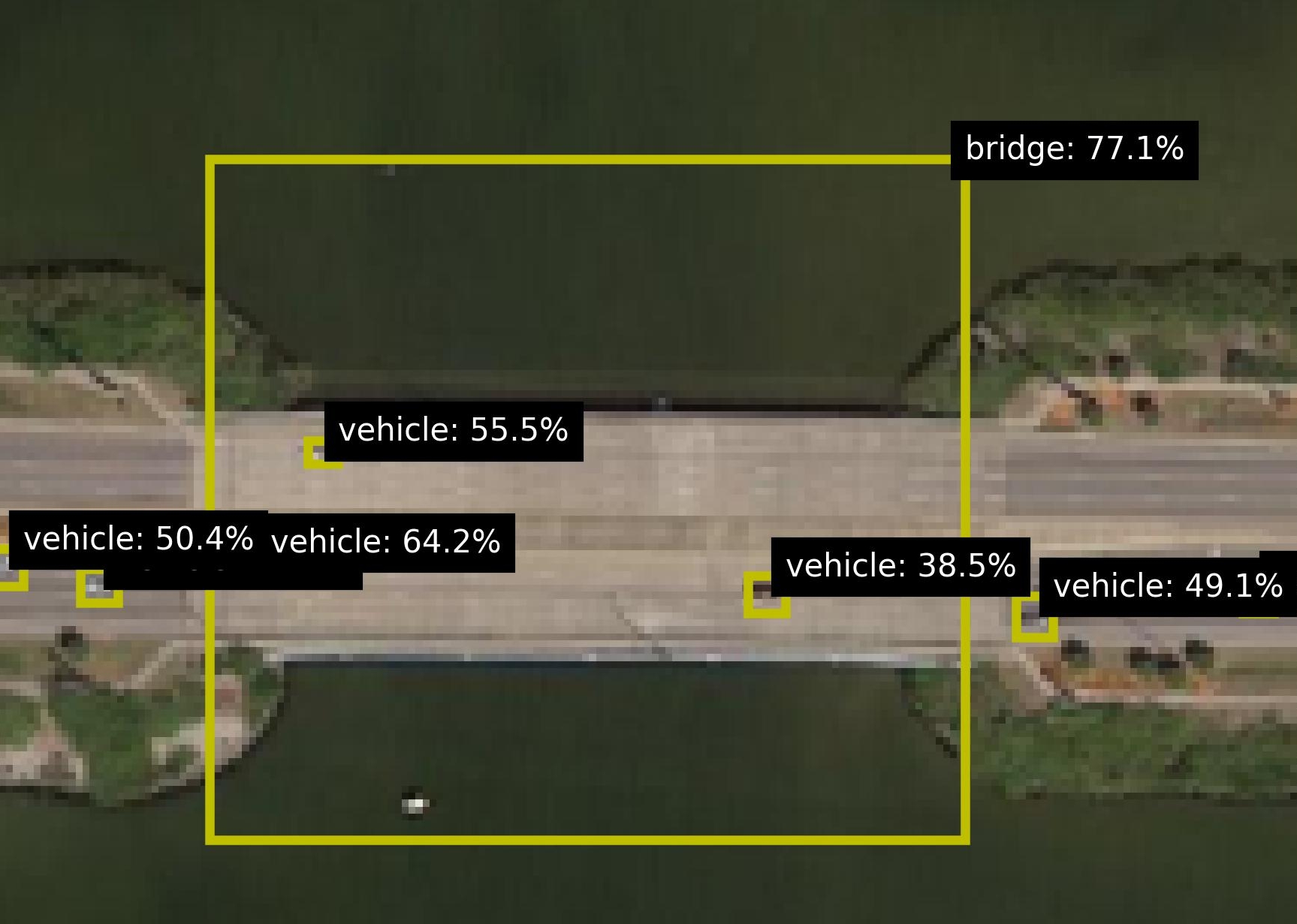}
         \caption{(zoom)}
         \label{pos_img_zoom}
     \end{subfigure}
     \begin{subfigure}[b]{0.5\textwidth}
         \centering
         \includegraphics[width=\textwidth]{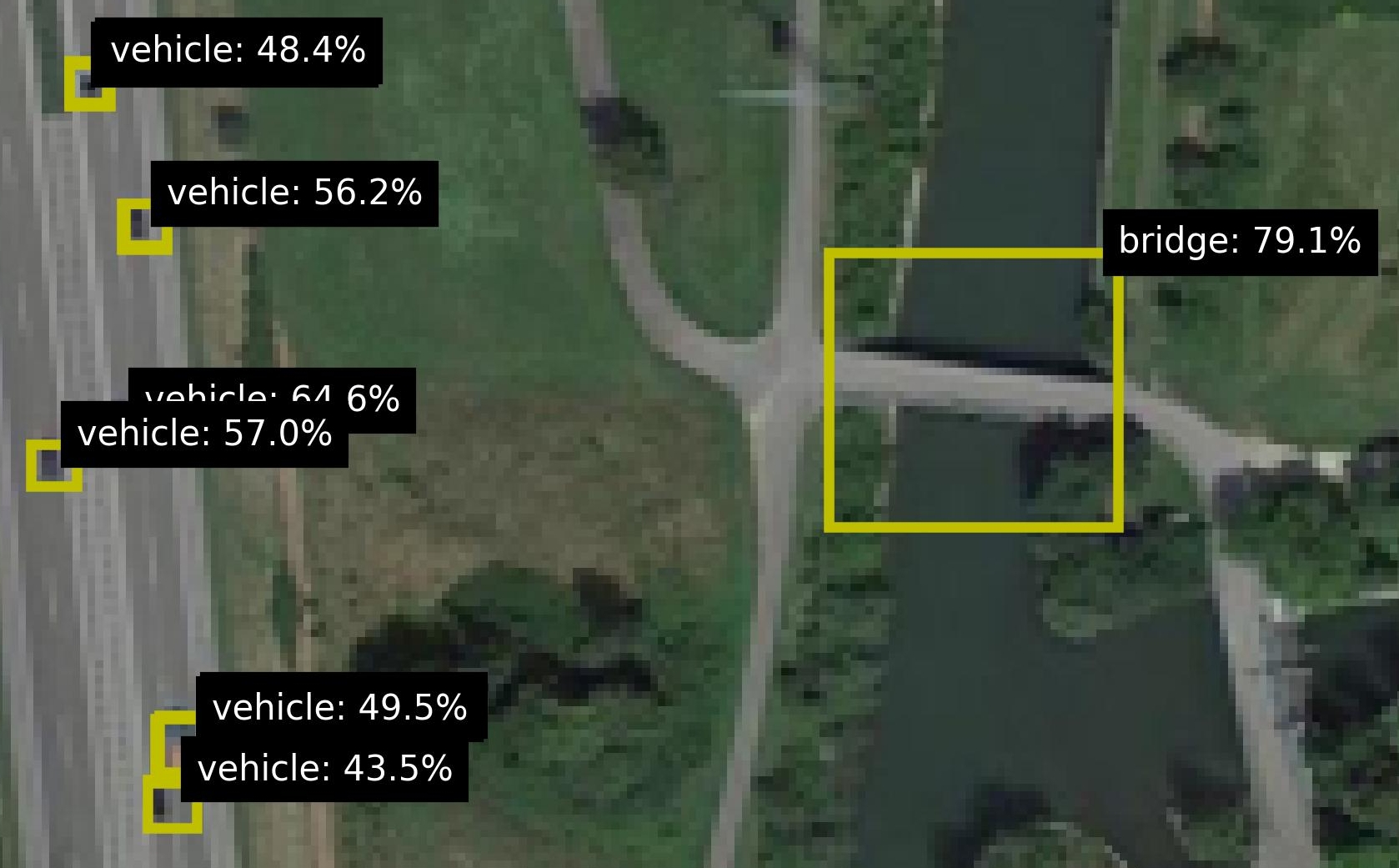}
         \caption{(zoom)}
         \label{neg_img_zoom}
     \end{subfigure}
     \caption{Examples of images with the detected objects and their probabilities.}
     \label{pos_neg_img}
\end{figure}

\begin{figure}[H]
    \centering
     \begin{subfigure}[b]{0.99\textwidth}
         \centering
         \includegraphics[width=0.31\textwidth]{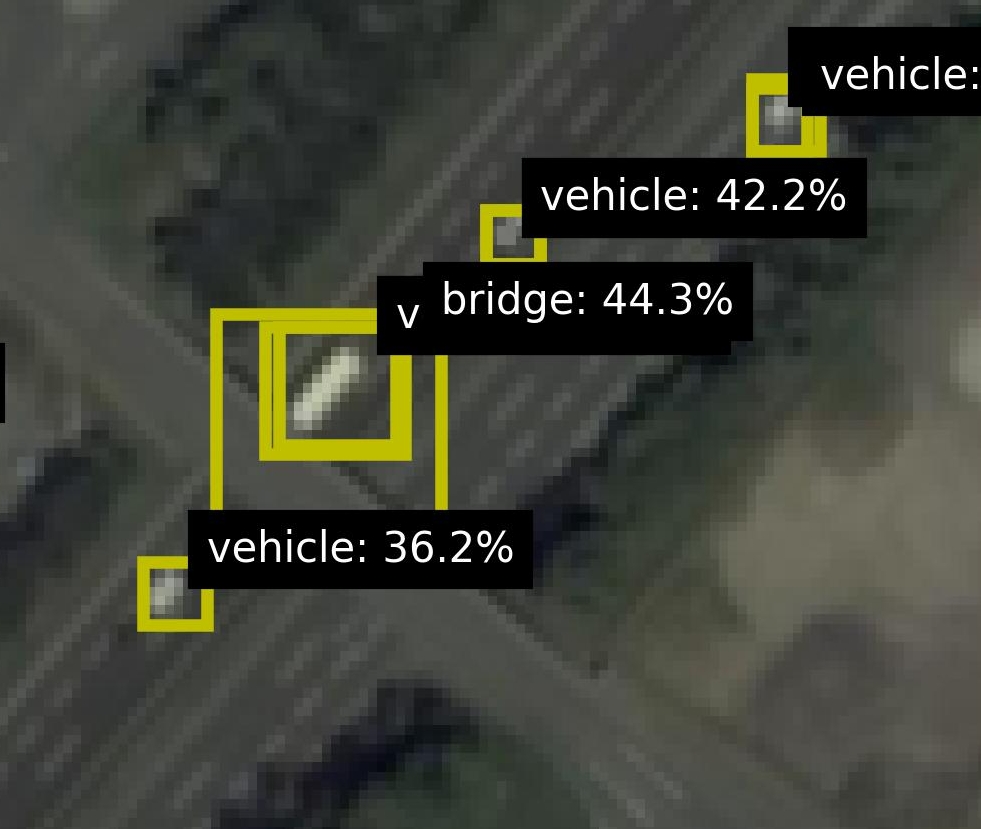}
         \centering
         \includegraphics[width=0.31\textwidth]{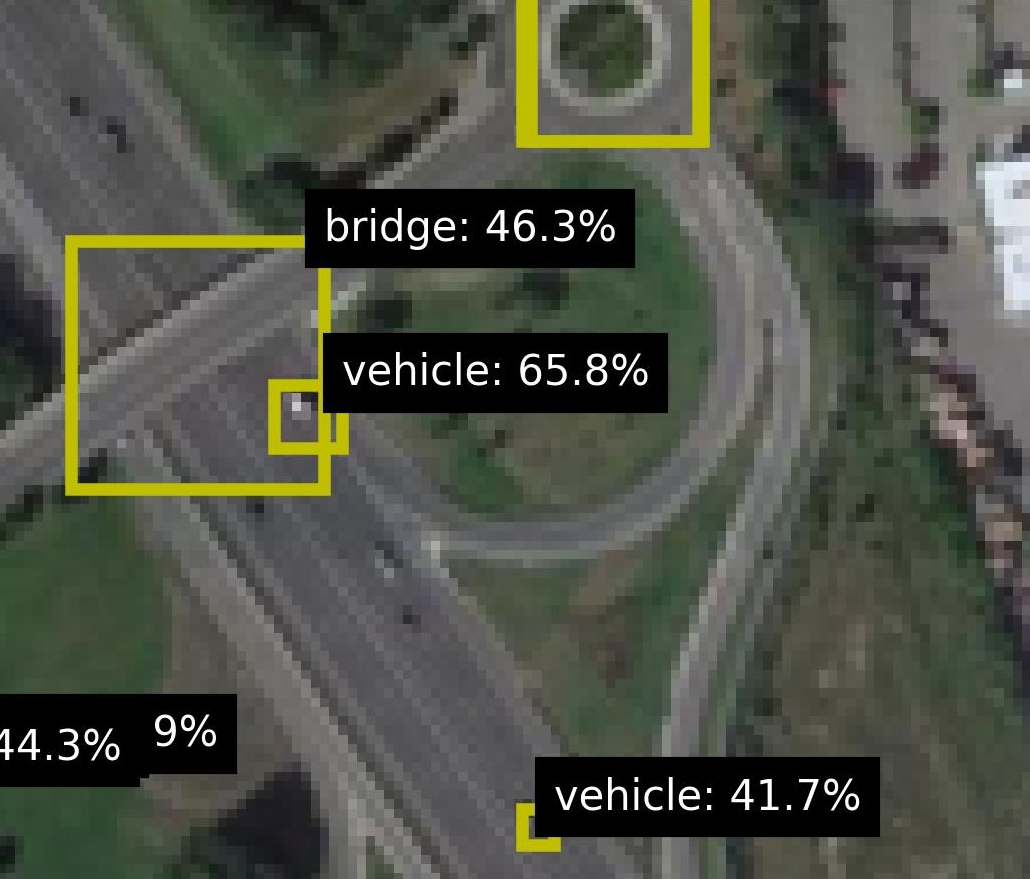}
         \centering
         \includegraphics[width=0.35\textwidth]{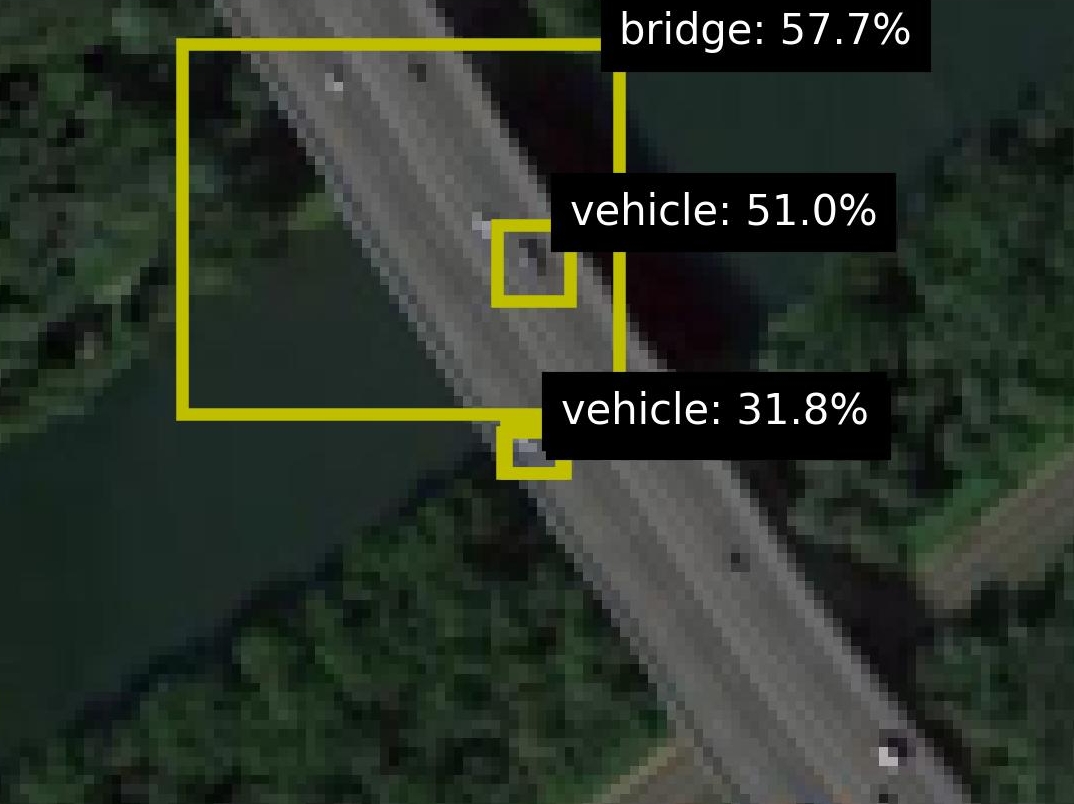}
     \end{subfigure}
     \caption{Hard cases due to incorrect groundtruths (right) or incorrect detections (others).}
     \label{hard_cases}
\end{figure}

For the background knowledge, we leverage the pretrained DOTA Aerial Images Model \autocite{dota_model} to predict the objects in the images, with for each object a label, location (bounding box) and a probability (confidence value). For each image, the respective predictions are added to the background knowledge, as a predicate with a confidence, e.g. 0.7 :: vehicle(x). The locations of the objects are used to calculate a confidence for two relations: \texttt{is\_on} and \texttt{is\_close}. This information is added to the background knowledge as well. Figure \ref{pos_neg_img} shows various images from the dataset, including zoomed versions to reveal some more details and to highlight the small size of the objects. Figure \ref{neg_img} shows an image with many objects. The relational patterns of interest is `vehicle on bridge'. For this pattern, there are 11 positive test images and 297 negative test images. Figure \ref{pos_neg_img} shows both a positive (left) and negative image (right). To make the task realistic, both sets contain images with vehicles, bridges and roundabouts, so the model cannot distinguish the positives and negatives by purely finding the right sets of objects; the model really needs to find the right pattern between the right objects. Out of the negative images, 17 are designated as hard, due to incorrect groundtruths (2 images) and incorrect detections (15 images). These hard cases are shown in Figure \ref{hard_cases}. 

\subsection{Experimental Setup}\label{experimental_setup}

The dataset is categorized into three subsets that are increasingly harder in terms of flaws in the background knowledge. Easy: This smallest subset excludes the incorrect groundtruths, a manual check that most object predictions are reasonable, i.e. images with many predicted objects are withheld (this includes images with many false positives). Intermediate: This subset excludes the incorrect groundtruths. Compared to Easy, this subset adds all images with many object predictions. Hard: This is the full set, which includes all images, also the ones with incorrect groundtruths. We are curious whether ILP methods can indeed generalize from small numbers of examples, as is hypothesized \autocite{cropper2022inductive}. Many datasets used in ILP are using training data with tens to hundreds (sometimes thousands) of labeled samples \autocite{hocquette2024learning,bellodi2015structure}. We investigate the performance for as few as \{1, 2, 4, 8\} labels for respectively the positive and negative set, as this is common in practical settings. Moreover, common ILP datasets are about binary background knowledge, without associated probabilities \autocite{hocquette2024learning,bellodi2015structure}. In contrast, we consider probabilistic background knowledge. From the Easy subset we construct an Easy-1.0 set by thresholding the background knowledge with a manually chosen optimal threshold, which results in an almost noiseless dataset and shows the complexity of the logical rule to learn. All experiments are repeated 5 times, randomly selecting the training samples from the dataset and using the rest of the data set as test set. 

\subsection{Model Variants and Baselines}

We compare Propper with Popper (on which it builds), to validate the merit of integrating the neurosymbolic inference and the continuous cost function BCE. Moreover, we compare these ILP models with statistical ML models: the Support Vector Machine \autocite{cortes1995support} (SVM) because it is used so often in practice; a Graph Neural Network \autocite{wu2020comprehensive} (GNN) because it is also relational by design which makes it a reasonable candidate for the task at hand i.e. finding a relational pattern between objects. All methods except the SVM are relational and permutation invariant. The objects are unordered and the models should therefore represent them in an orderless manner. The SVM is not permutation invariant, as objects and their features have some arbitrary but designated position in its feature vectors. All methods except Popper are probabilistic. All methods except the most basic Popper variant, can handle some degree of noise. The expected noise level for NoisyCombo is set at 0.15. The tested models are characterized in Table \ref{methods_baselines}. 

\begin{table}[H]
 \centering
 \caption{The tested model variants and their properties.}\label{methods_baselines}
 {\tablefont\begin{tabular}{@{\extracolsep{\fill}}llllll}
   \topline
    Model & 
    SVM &
    GNN &
    ILP Popper &
    ILP MaxSynth &
    ILP Propper \\
    & 
    Cortes 1995 & 
    Wu 2020 & 
    Cropper 2021 & 
    Hoguette 2024 & 
    (ours)
    \midline
    Constrainer   & - & - & Combo & MaxSynth & Noisy-Combo \\  
    Tester        & - & - & Prolog & Prolog & Scallop \\  
    Cost function & - & - & MDL & MDL & BCE 
    \midline
    Type & Stat. & Stat. & Logic & Logic & Logic \\  
    Label noise & Yes & Yes & No & Yes & Yes \\
    Background noise & Yes & Yes & No & Some & Yes \\
    Relational & No & Yes & Yes & Yes & Yes \\
    Permutation inv. & No & Yes & Yes & Yes & Yes \\ 
    Probabilistic & Yes & Yes & No & No & Yes 
   \botline
    \end{tabular}}
\end{table}

For a valid comparison, we increase the SVM's robustness against arbitrary object order. With prior knowledge about the relevant objects for the pattern at hand, these objects can be placed in front of the feature vector. This preprocessing step makes the SVM model less dependent on the arbitrary order of objects. In the remainder of the analyses, we call this variant `SVM ordered'. To binarize the probabilistic background knowledge as input for Popper, the detections are thresholded with the general value of 0.5. 

\subsection{Increasing Noise in Background Knowledge}

\begin{figure}[H]
    \centering
     \includegraphics[width=0.98\textwidth]{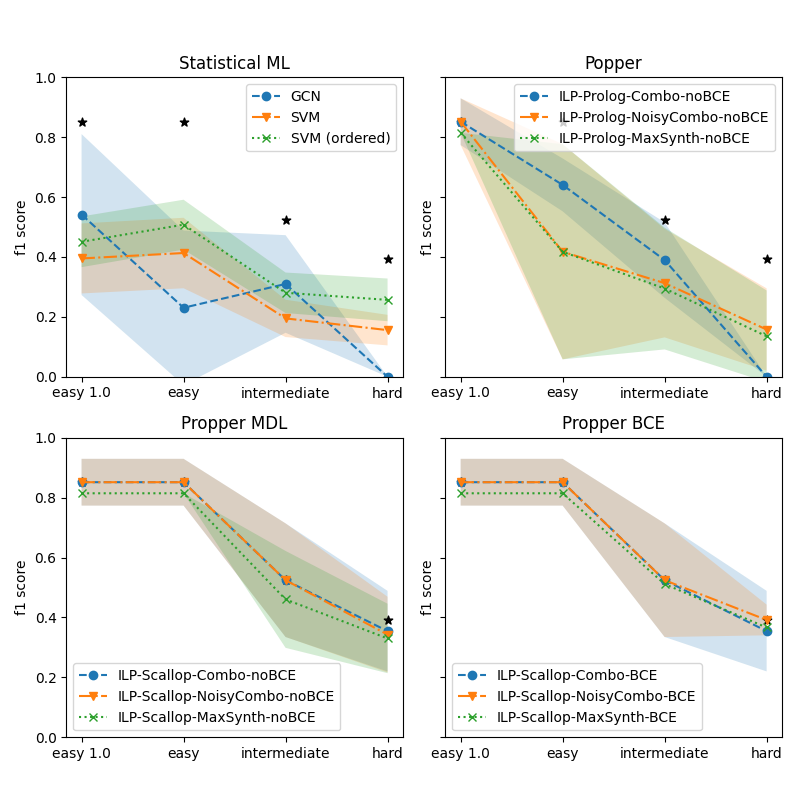}
     \caption{Performance of the models on finding a relational pattern in satellite images, for increasing hardness of image sets. The best performer is Propper BCE, indicated in each graph by * for comparison. Our probabilistic ILP outperforms binary ILP and statistical ML.}
     \label{performance_hardness}
\end{figure}

We are interested in how the robustness of model learning for increasing difficulty of the dataset. Here we investigate the performance on the three subsets from Section \ref{experimental_setup}: Easy, Intermediate and Hard. Figure \ref{performance_hardness} shows the performance for various models for increasing difficulty. The four subplots show the various types of models. For a reference, the best performing model is indicated by an asterisk (*) in all subplots. It is clear that for increasing difficulty, all models struggle. The statistical ML models struggle the most: the performance of the GNN drops to zero on the Hard set. The SVMs are a bit more robust but the performance on the Hard set is very low. The most basic variant of Popper also drops to zero. The noise-tolerant Popper variants (Noisy-Combo and MaxSynth) perform similarly to the SVMs. Propper outperforms all models. This finding holds for all Propper variants (Combo, Noisy-Combo and MaxSynth). Using BCE as a cost function yields a small but negligible advantage over MDL.

\subsection{Learning Efficiency with Few Labels}

\begin{figure}[H]
    \centering
     \includegraphics[width=0.95\textwidth]{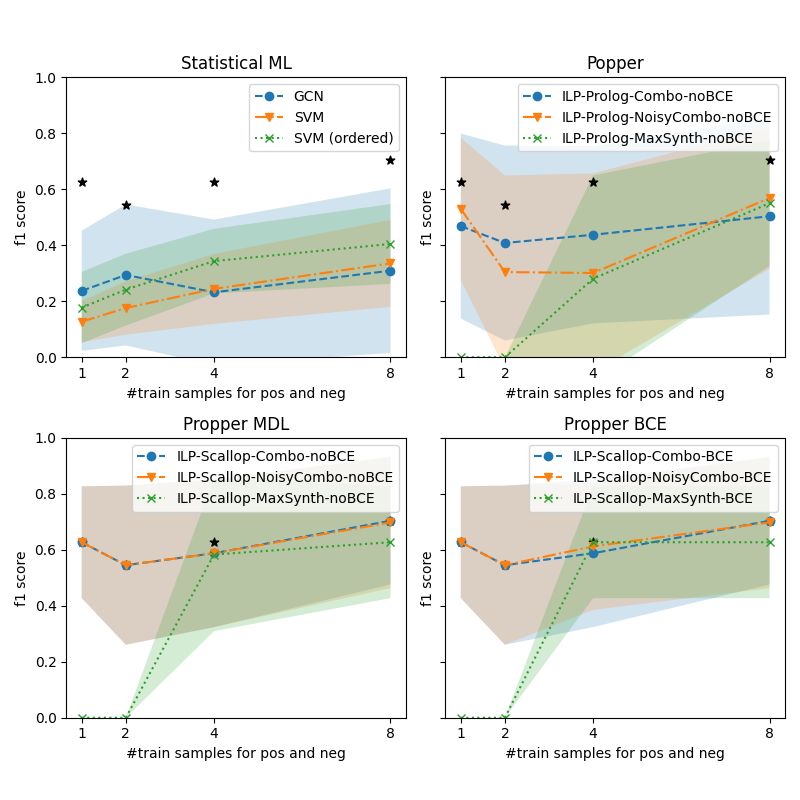}
     \caption{Performance of the models on finding a relational pattern in satellite images, for increasing training sets. The best performer is Propper BCE, indicated in each graph by * for comparison. Our probabilistic ILP outperforms binary ILP and statistical ML.}
     \label{performance_samples}
\end{figure}

We are curious how the models perform with as few as \{1, 2, 4, 8\} labels for respectively the positive and negative set. The performance is measured on the Hard set. Figure \ref{performance_samples} shows the performance for various models for increasing training set size. The four subplots show the various types of models. Again, for reference, the best performing model is indicated by an asterisk (*) in all subplots. The upper left shows the statistical ML models. They do perform better with more training samples, but the performance is inferior to the ILP model variants. The Propper variant with Scallop and Noisy-Combo and BCE is the best performer. BCE does not improve significantly over MDL. MaxSynth has an optimization criterion that cannot operate with less than three training samples. The main improvement by Propper is observed when switching from Combo to Noisy-Combo and switching from Prolog to Scallop (i.e. neurosymbolic inference).

\subsection{Second Dataset}

\begin{figure}[H]
    \centering
    \begin{subfigure}[b]{0.99\textwidth}
    \centering
    \includegraphics[width=0.46\textwidth]{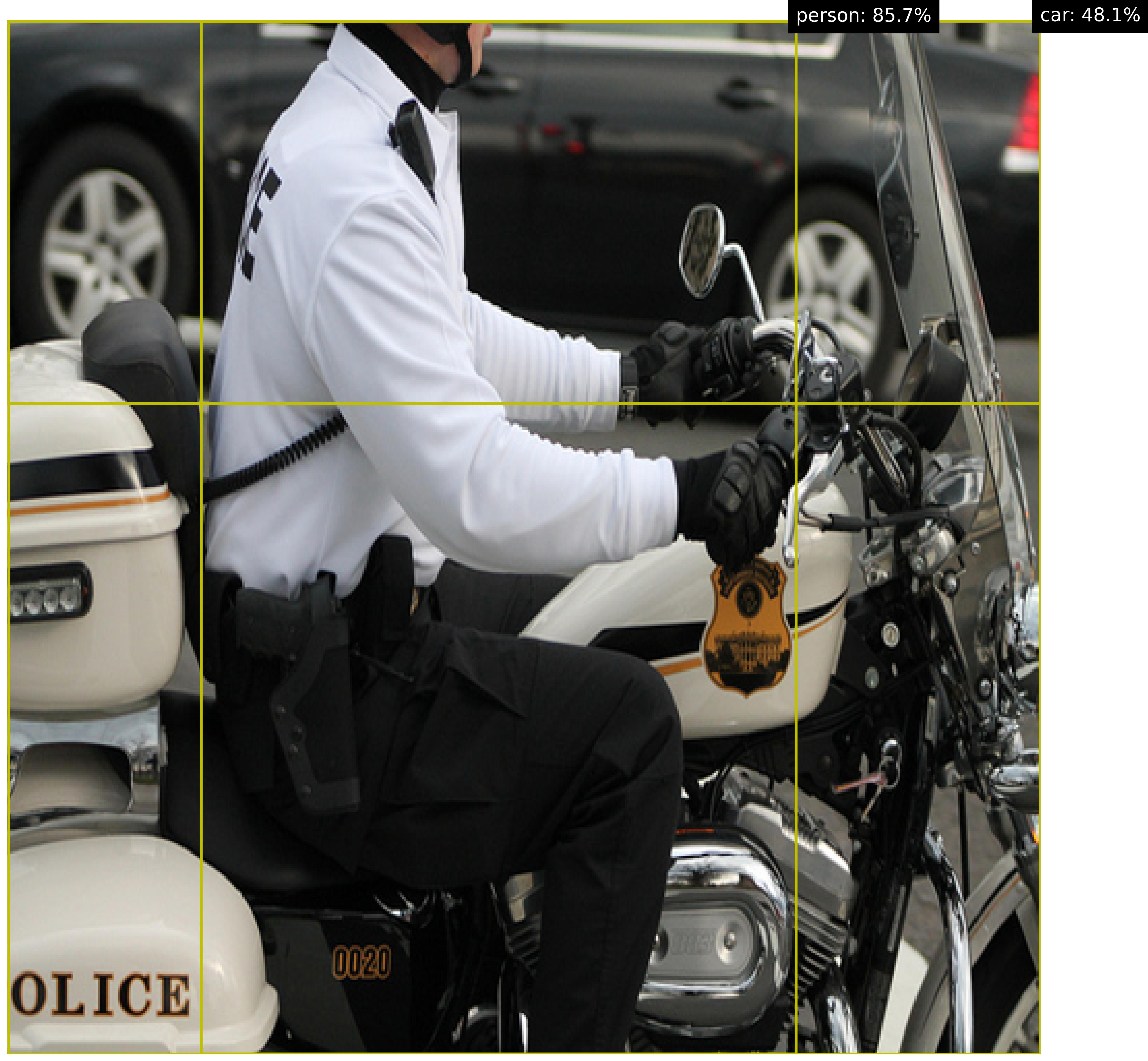}
    \includegraphics[width=0.44\textwidth]{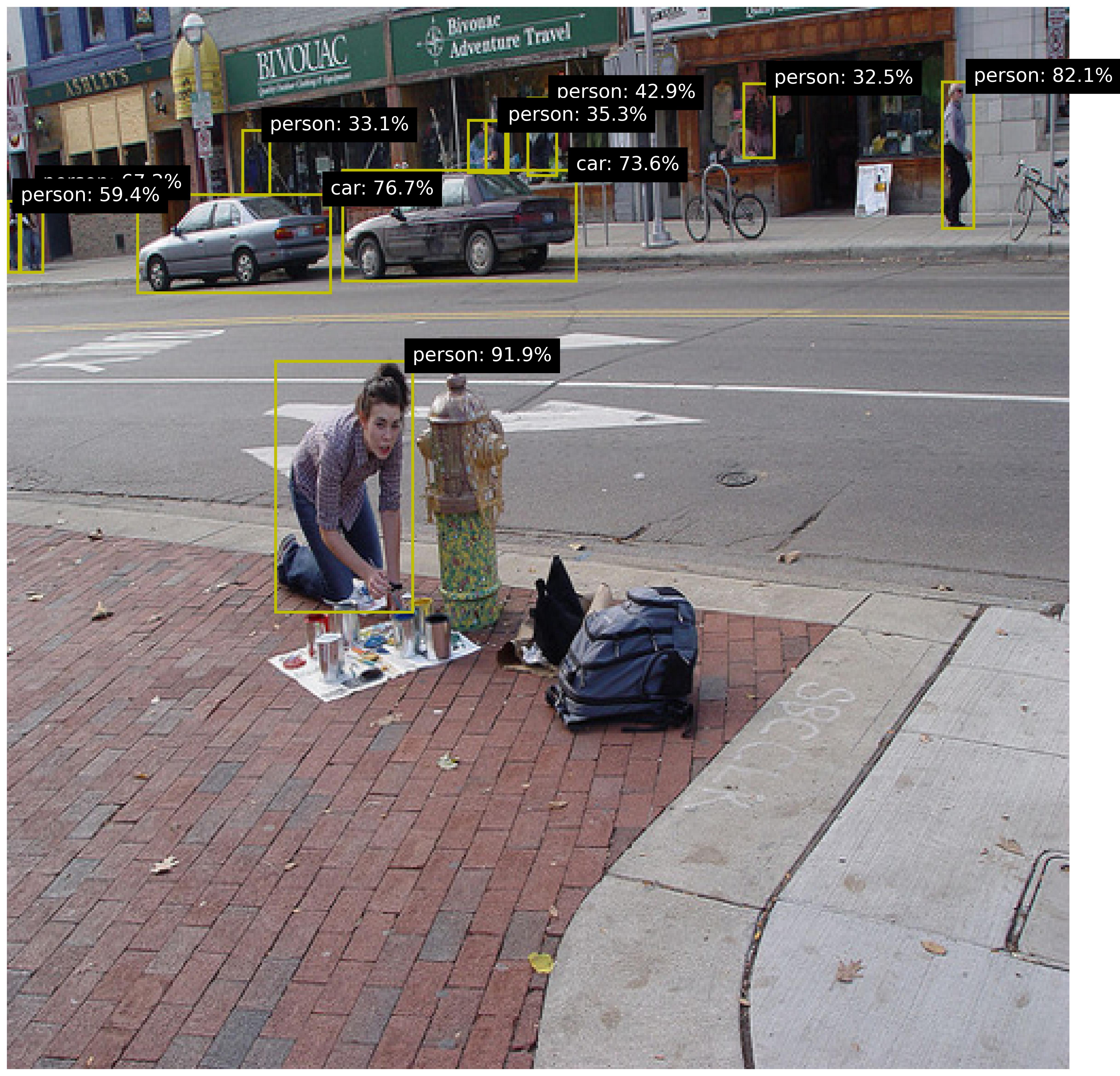}
     \end{subfigure}
     \caption{Examples of the MS-COCO dataset with images of everyday scenes.}
     \label{coco_imgs}
\end{figure}

We are interested how the methods perform on a different dataset. The MS-COCO dataset \autocite{lin2014microsoft} contains a broad variety of images of everyday scenes. This dataset is challenging, because there are many different objects in a wide range of settings. Similar to the previous experiment, the background knowledge is acquired by the predictions of a pretrained model, GroundingDINO \autocite{liu2023groundingDINO}, which are used to extract the same two relations. Figure \ref{coco_imgs} shows some examples.

The pattern of interest is `person next to a car'. We consider all images that have a maximum of two persons and two cars, yielding 1728 images. We use random 8 positive and 8 negative images for training, which is repeated 5 times. We test both ILP variants, Popper and Propper, for the MaxSynth constrainer, because the Combo constrainer regularly did not return a solution. We validate Popper with various thresholds to be included as background knowledge. Propper does not need such a threshold beforehand, as all background knowledge is considered in a probabilistic manner. The results are shown in Table \ref{coco}. Propper is the best performer, achieving f1 = 0.947. This is significantly better than the alternatives: SVM achieves f1 = 0.668 (-0.279) and Popper achieves f1 = 0.596 (-0.351). Adding probabilistic behavior to ILP is helpful for challenging datasets.

\begin{table}[H]
 \centering
 \caption{Model variants and performance on MS-COCO.}\label{coco}
 {\tablefont\begin{tabular}{@{\extracolsep{\fill}}lllllll}
   \topline
    category & method & constrainer & tester & criterion & threshold & f1 
   \midline
    ILP & Propper (ours) & MaxSynth & probabilistic & MDL & - & 0.947 \\
    ILP & Propper (ours) & MaxSynth & probabilistic & BCE & - & 0.754 \\
    Statistical ML & SVM & - & - & - & - & 0.668 \\
    Statistical ML & SVM (ordered) &  - & - & - & - & 0.652 \\
    ILP & Popper & MaxSynth & Prolog & MDL & 0.3 & 0.596 \\
    ILP & Popper & MaxSynth & Prolog & MDL & 0.5 & 0.466 \\
    ILP & Popper & MaxSynth & Prolog & MDL & 0.4 & 0.320 
   \botline
   \end{tabular}}
\end{table}

Table \ref{learned_programs} shows the learned programs, how often each program was predicted across the experimental repetitions, and the respective resulting f1 scores. The best program is that there is a person on a car. Popper yields the same program, however, with a lower f1-score, since the background knowledge is thresholded before learning the program, removing important data from the background knowledge. This confirms that in practice it is intractable to set a perfect threshold on the background knowledge. It is beneficial to use Propper which avoids such prior thresholding. 

\begin{table}[H]
 \centering
 \caption{Learned programs, prevalence and performance on MS-COCO.}\label{learned_programs}
 \vspace{-0.5cm}
 {\tablefont\begin{tabular}{@{\extracolsep{\fill}}llclllll}
   \topline
    model & f1 & \% & \multicolumn{2}{l}{program} &  &  & 
   \midline
    Propper & 1.0 & 40 & \multicolumn{5}{l}{\texttt{f(A) :- has\_object(A,B), person(B), is\_on(B,C), car(C).}} \\
    Propper & 0.93 & 40 & \multicolumn{5}{l}{\texttt{f(A) :- has\_object(A,B), car(B), is\_on(C,B).}} \\
    Propper & 0.88 & 20 & \multicolumn{5}{l}{\texttt{f(A) :- person(B), has\_object(A,C), car(B).}} 
    \midline
    Popper & 0.82 & 20 & \multicolumn{5}{l}{\texttt{f(A) :- has\_object(A,C), is\_on(C,B), has\_object(A,B).}} \\
    Popper & 0.72 & 20 & \multicolumn{5}{l}{\texttt{f(A) :- has\_object(A,C), is\_on(B,C), person(B).}} \\
    Popper & 0.72 & 40 & \multicolumn{5}{l}{\texttt{f(A) :- person(C), is\_on(C,B), has\_object(A,C), car(B).}} \\
    Popper & 0 & 20 & \multicolumn{5}{l}{\texttt{No program learned.}}
   \botline
   \end{tabular}}
\end{table}

\vspace{-1.2cm}

\section{Discussion and Conclusions}

We proposed Propper, which handles flawed and probabilistic background knowledge by extending ILP with a combination of neurosymbolic inference, a continuous criterion for hypothesis selection (BCE), and a relaxation of the hypothesis constrainer (NoisyCombo). Neurosymbolic inference has a significant impact on the results. Its advantage is that it does not need prior thresholding on the probabilistic background knowledge (BK), which is needed for binary ILP and is always imperfect. NoisyCombo has a small yet positive effect. It provides a parameter for the level of noise in BK, which can be tailored to the dataset at hand. The BCE has little impact. Propper is able to learn a logic program about a relational pattern that distinguishes between two sets of images, even if the background knowledge is provided by an imperfect neural network that predicts concepts in the images with some confidence. With as few as a handful of examples, Propper learns effective programs and outperforms statistical ML methods such as a GNN. 

Although we evaluated Propper on two common datasets with different recording conditions, a broader evaluation of Propper across various domains and datasets to confirm its generalizability and robustness for various (especially non-image) use cases, is interesting. 
The proposed framework of integrated components allows for an easy setup of the system and simple adaptation to new developments/algorithms within the separate components. However, the integration as is performed now could be non-optimal in terms of computational efficiency. For example the output of the hypothesis generation is an answer set, which in Popper is converted to Prolog syntax. Propper converts this Prolog syntax to Scallop syntax. Developing a direct conversion from the answer sets to the Scallop syntax is recommended. We favored modularization over full integration and computational efficiency, in order to facilitate the methodological configuration and comparison of the various components. It is interesting to investigate whether a redesign of the whole system with the components integrated will lead to a better system. To make the step to fully probabilistic ILP, the allowance of probabilistic rules should be added to the system as well, for example by integration of StarAI methods \autocite{raedt2016statistical}. 



\printbibliography

\end{document}